\documentclass[runningheads]{llncs}
\usepackage[T1]{fontenc}

\usepackage{amsmath}
\usepackage{hyperref}
\usepackage{booktabs}
\usepackage{amsfonts}
\usepackage{subcaption}
\usepackage{orcidlink}

\usepackage{graphicx,verbatim}

\usepackage{color}

\begin{document}
\title{Predicting Postprandial Glycemic Response from Meal Images, Clinical Variables, and Gut Microbiome Information}
\titlerunning{Multimodal PPGR Prediction}

\author{
Varvara Kondratyeva\inst{1,2}\orcidlink{0009-0008-6997-9006} \and
Kamilia Zaripova\inst{1,3}\orcidlink{0000-0002-8200-0818} \and \\
Nassir Navab\inst{1,3}\orcidlink{0000-0002-6032-5611} \and
Azade Farshad\inst{1,3,4}\orcidlink{0000-0002-1080-1587}
}

\authorrunning{V. Kondratyeva et al.}

\institute{
Technical University of Munich, Munich, Germany \and
University Hospital Augsburg, Augsburg, Germany \and
Munich Center for Machine Learning (MCML), Munich, Germany \and
Aalto University, Espoo, Finland \\
\email{varvara.kondratyeva@tum.de}
}

\maketitle              

\begin{abstract}
Predicting postprandial glycemic response (PPGR) is fundamental to personalized nutrition and type~2 diabetes management, yet existing approaches typically rely on manually reported dietary intake, limiting their scalability in free-living settings. We propose a multimodal framework that replaces manual dietary logging with image-derived macronutrient estimates and integrates them with clinical variables and gut microbiome information for personalized PPGR prediction. The framework jointly performs image-based macronutrient estimation and glucose prediction, while an attention-based prediction module models interactions between dietary and host-specific information. We evaluate the proposed approach on a real-world dataset comprising meal images, continuous glucose monitoring, clinical variables, and gut microbiome profiles. The proposed model outperforms existing PPGR baselines using image-derived nutritional inputs and approaches the performance of methods that rely on manually reported macronutrients despite using automatically estimated nutritional information. These results demonstrate that combining image-derived nutrition with complementary clinical and gut microbiome information provides a practical foundation for scalable personalized PPGR prediction.
\end{abstract}


\section{Introduction}

Predicting postprandial glycemic response (PPGR), the rise in blood glucose following a meal, is a fundamental problem in personalized nutrition and Type~2 diabetes (T2D) management. Unlike a single glucose measurement, PPGR reflects both the magnitude and duration of glucose excursions, making summary measures such as the incremental area under the glucose curve over 120 minutes (iAUC120) more informative for assessing glycemic response. Elevated postprandial glucose is associated with metabolic and cardiovascular risk and may precede abnormalities in fasting glucose, making meal-response prediction relevant for early intervention \cite{zeevi_personalized_2015,bergman_international_2024}. However, PPGR varies substantially across individuals, even for identical meals due to differences in physiology, lifestyle, medication use, and gut microbiome composition \cite{berry_human_2020,cavalot_postprandial_2011,mendes-soares_model_2019}, motivating personalized multimodal prediction models.

Recent machine learning approaches combine dietary information with clinical and microbiome data to improve PPGR prediction \cite{zeevi_personalized_2015,berry_human_2020,zhang_joint_2023,popova_personalized_2025}. However, they typically rely on manually reported or expert-annotated macronutrients, which are time-consuming to collect and prone to recall and portion-size estimation errors. Computer vision models can estimate nutritional properties directly from meal images \cite{thames_nutrition5k_2021}, enabling passive dietary assessment in free-living settings. Whether such image-derived nutritional estimates can replace manually reported dietary inputs for personalized PPGR prediction remains largely unexplored.

This problem presents several machine learning challenges. Image-derived macronutrients are inherently noisy, while PPGR depends on complex interactions between meal composition and host-specific factors. Learning these relationships from heterogeneous visual, clinical, and microbiome data is further complicated by substantial inter-individual variability and the limited availability of multimodal datasets.

In this work, we propose a multimodal framework for personalized PPGR prediction that combines image-derived macronutrient estimates with clinical variables and gut microbiome information, eliminating the need for manual dietary logging. We further investigate end-to-end learning of image-based macronutrient estimation and PPGR prediction within a multimodal learning framework. We evaluate different strategies for integrating image-derived dietary, clinical, and gut microbiome information for personalized PPGR prediction. 

Our contributions are summarized as follows:
\begin{itemize}
\item We propose a multimodal framework for PPGR prediction that integrates image-derived macronutrient estimates, clinical variables, and gut microbiome information without requiring manual dietary logging.
\item We investigate end-to-end learning of image-based macronutrient estimation and PPGR prediction within a unified framework.
\item We provide a comprehensive empirical evaluation of multimodal fusion strategies for integrating image-derived dietary, clinical, and gut microbiome information for personalized PPGR prediction.
\item We demonstrate that image-derived macronutrients achieve competitive PPGR prediction compared with manually reported nutritional inputs under free-living conditions.
\end{itemize}
\section{Related Work}

\noindent\textbf{Personalized PPGR prediction.}
Personalized PPGR prediction has been extensively studied using machine learning models that integrate dietary, clinical, and microbiome information \cite{zeevi_personalized_2015,berry_human_2020,popova_personalized_2025,metwally_use_2025}. These studies consistently show that incorporating host-specific information improves prediction accuracy but rely on manually reported or expert-annotated nutritional inputs. Moreover, these studies treat PPGR prediction as a supervised tabular regression task using tree-based models such as RandomForestRegressor \cite{berry_human_2020} or LightGBM \cite{popova_personalized_2025}.

\noindent\textbf{Vision-based dietary assessment.}
Recent computer vision methods estimate calories and macronutrients directly from food images \cite{thames_nutrition5k_2021,shao_vision-based_2023}. Several multimodal approaches further combine meal images with physiological signals, such as Continuous Glucose Monitoring (CGM), to improve nutritional estimation \cite{zhang_joint_2023,das_predicting_2022,sajjadi_towards_2021}. However, these methods focus on dietary assessment rather than predicting physiological responses.

\noindent\textbf{Multimodal glucose prediction.} Khan \emph{et al.} \cite{khan_automatic_2022} demonstrated that food images can predict meal glycemic impact, while Jaggi \emph{et al.} \cite{jaggi_automated_2025} incorporated meal photographs into glucose response prediction for Type 1 diabetes. Nevertheless, multimodal PPGR prediction in Type 2 diabetes remains largely unexplored.

Existing work demonstrates that nutritional information can be estimated from food images and that multimodal data improve PPGR prediction. However, whether image-estimated macronutrients can replace manually reported nutritional inputs has not been systematically investigated. 
Progress is further limited by the scarcity of datasets jointly containing food images, CGM, clinical variables, and gut microbiome profiles. Collecting these modalities in the same participants requires synchronized meal imaging and CGM acquisition, clinical phenotyping, and microbiome sequencing, making such studies costly, logistically demanding, and prone to missing data. As a result, complex multimodal models remain difficult to train and evaluate at scale. Here, we leverage a dataset containing these complementary modalities to study personalized PPGR prediction from image-estimated macronutrients, clinical variables, and microbiome profiles.

\section{Methods}
\begin{figure}[t]
    \centering
    \includegraphics[height=3.5cm]{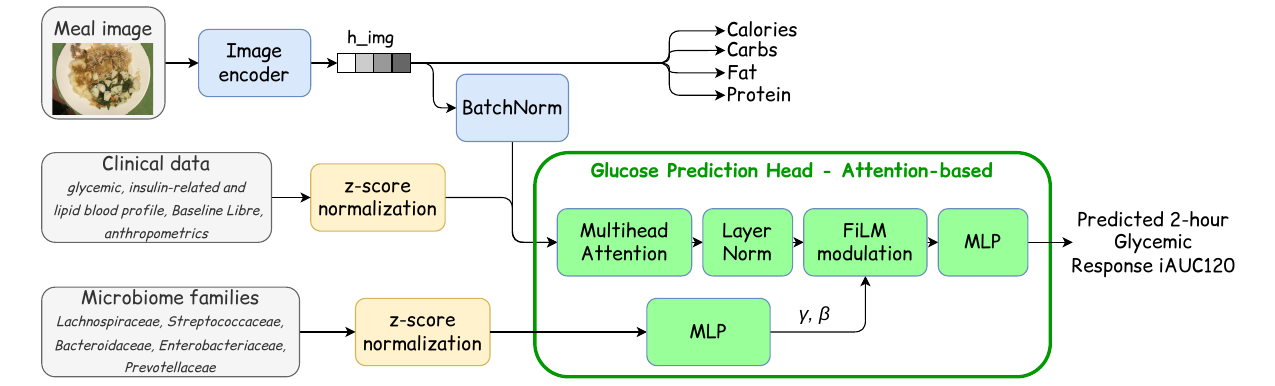}
    \caption{Overview of the proposed multimodal framework for personalized PPGR prediction. A meal image is processed by the image encoder to produce image-derived macronutrient estimates, which are fused with clinical features using multi-head self-attention. Gut microbiome features condition the fused representation through FiLM modulation before the regression head jointly predicts $\mathrm{iAUC120}$ and the auxiliary target $\mathrm{AUC120}$.}
    \label{fig:architecture}
\end{figure}

\subsection{Problem Formulation}

We propose a multimodal model that fuses meal photographs, clinical data, and gut microbiome composition to predict postprandial glycemic response (PPGR) in patients with type~2 diabetes/prediabetes.

PPGR is quantified using the incremental area under the glucose curve over 120 minutes (iAUC120), which measures the cumulative increase in blood glucose above the pre-meal baseline during the first two hours after meal consumption. Unlike a single glucose measurement, iAUC120 summarizes both the magnitude and duration of postprandial glucose elevation and therefore provides a comprehensive measure of an individual's glycemic response to a meal.

Given an RGB meal image $\mathbf{x}_{\mathrm{rgb}}$, clinical features $\mathbf{x}_{\mathrm{clin}}$, and microbiome features $\mathbf{x}_{\mathrm{micro}}$, the model predicts the corresponding iAUC120 value:

\begin{equation}
\hat{y}_i = f_{\theta}\!\left(\mathbf{x}^{(i)}_{\mathrm{rgb}}, \mathbf{x}^{(i)}_{\mathrm{clin}}, \mathbf{x}^{(i)}_{\mathrm{micro}}\right).
\label{eq:sample_model}
\end{equation}

During training, absolute area under the glucose curve over 120 minutes ($\mathrm{AUC120}$) is used as an auxiliary prediction target to regularize learning.

The model $f_\theta$ consists of an image encoder $h_{\mathrm{img}}$, which maps the meal image to a macronutrient-aligned embedding $\mathbf{h}_{\mathrm{img}}$, and a glucose prediction head $h_{\mathrm{glucose}}$, which combines this representation with clinical and microbiome features:

\begin{equation}
\hat{y} =
h_{\mathrm{glucose}}
\!\left(
h_{\mathrm{img}}(\mathbf{x}_{\mathrm{rgb}}),
\mathbf{x}_{\mathrm{clin}},
\mathbf{x}_{\mathrm{micro}}
\right).
\end{equation}
 
\subsection{Image Encoder}
 
We replicate the macronutrient-estimation model of Shao et al.~\cite{shao_vision-based_2023}: a two-branch ResNet-101 backbone, pretrained on Food2K and fine-tuned on Nutrition5K, predicting calories, fat, carbohydrate, and protein from RGB(-D) input. As depth is unavailable in our free-living setting, the depth branch is replaced with a duplicate RGB image. We constrain $\mathbf{h}_{\mathrm{img}} \in \mathbb{R}^4$ so that each dimension corresponds directly to one macronutrient, making the macronutrient decoder an identity mapping,
\begin{equation}
\hat{\mathbf{x}}_{\mathrm{macro}} = g_{\mathrm{macro}}(\mathbf{h}_{\mathrm{img}}) = \mathbf{h}_{\mathrm{img}},
\end{equation}
and the latent space directly interpretable as macronutrient estimates.
 
\subsection{Attention-Based Glucose Prediction Head}

Our proposed glucose prediction head models interactions between image-derived macronutrient and clinical features using multi-head self-attention while conditioning the fused representation on gut microbiome information through Feature-wise Linear Modulation (FiLM). To this end, the macronutrient and clinical features are embedded as tokens and processed with multi-head self-attention (MHA):

\begin{equation}
\tilde{T} = \mathrm{LayerNorm}\!\left(T + \mathrm{MHA}(T,T,T)\right).
\end{equation}

Rather than being concatenated directly, microbiome features are mapped by a small MLP to FiLM parameters $(\gamma,\beta)$ \cite{perez_film_2017}, which modulate the attended token representations (Fig.~\ref{fig:architecture}):

\begin{equation}
T^{*}_i = \tilde{T}_i \odot (1+\gamma) + \beta.
\end{equation}

The additive offset in the scaling term follows the residual FiLM parameterization, where centering the scale around one preserves the identity mapping when $\gamma=0$ and $\beta=0$ \cite{wisnu_stsm-film_2025}. The modulated tokens are flattened and passed to a two-layer MLP regression head to jointly predict $\log(\mathrm{iAUC120})$ and the auxiliary target $\log(\mathrm{AUC120})$. Dropout is applied in the attention, FiLM, and prediction modules. When microbiome data are unavailable, FiLM modulation is omitted and the attended token representation $\tilde{T}$ is passed directly to the regression head.
\subsection{Joint Learning}

The image encoder and glucose prediction head are trained end-to-end. During joint training, the predicted macronutrient vector is batch-normalized before fusion, while the clinical ($\tilde{\mathbf{x}}_{\mathrm{clin}}$) and microbiome ($\tilde{\mathbf{x}}_{\mathrm{micro}}$) features are z-score standardized:

\begin{equation}
\hat{y} =
h_{\mathrm{glucose}}\!\left(
\mathrm{BN}\!\left(h_{\mathrm{img}}(\mathbf{x}_{\mathrm{rgb}})\right),
\, \tilde{\mathbf{x}}_{\mathrm{clin}},
\, \tilde{\mathbf{x}}_{\mathrm{micro}}
\right).
\end{equation}

The network is jointly optimized to predict both $\log(\mathrm{iAUC120})$ and the auxiliary target $\log(\mathrm{AUC120})$.

\section{Results and Discussion}
\begin{table}[t]
\centering
\caption{Pearson correlation ($R$) between predicted and measured $\mathrm{iAUC120}$.
The upper block uses ground-truth (GT) macronutrients (oracle setting), while the lower block uses image-estimated macronutrients. Published results from Zeevi et al.~\cite{zeevi_personalized_2015} and Berry et al.~\cite{berry_human_2020} are shown for reference only.}
\small
\setlength{\tabcolsep}{4pt}
\begin{tabular}{lccc}
\toprule
\textbf{Model} & \textbf{R} & \textbf{p-value} & \textbf{95\% CI} \\
\midrule

\multicolumn{4}{l}{\textit{Oracle (GT macronutrients), original cohorts}} \\

Zeevi et al. & 0.68 & $<10^{-10}$ & -- \\
Berry et al. & 0.77 & -- & -- \\
\midrule
{\textit{Oracle (GT macronutrients), CGMacros dataset}} \\
Berry: Clinical + Macronutrients & 0.66 & $1.82{\times}10^{-8}$ & (0.49, 0.80) \\
Berry: Clinical + Macronutrients + Microbiome & 0.58 & $1.20{\times}10^{-6}$ & (0.42, 0.73) \\

\addlinespace[0.4mm]

Qwen3.5-4B: Clinical + Macronutrients & 0.50 & $5.58{\times}10^{-5}$ & (0.32, 0.67) \\
Qwen3.5-4B: Clin. + Macronutr. + Microbiome & 0.40 & $1.83{\times}10^{-3}$ & (0.16, 0.64) \\
Lingshu-7B: Clinical + Macronutrients & -0.03 & $8.00{\times}10^{-1}$ & (-0.09, 0.04) \\
Lingshu-7B: Clin. + Macronutr. + Microbiome & -0.28 & $2.88{\times}10^{-2}$ & (-0.61, -0.01) \\

\addlinespace[0.4mm]

MLP: Clinical + Macronutrients & 0.59 & $9.89{\times}10^{-7}$ & (0.49, 0.75) \\
MLP: Clinical + Macronutrients + Microbiome & 0.72 & $1.65{\times}10^{-10}$ & (0.58, 0.83) \\
Attention: Clinical + Macronutrients & 0.67 & $5.14{\times}10^{-9}$ & (0.53, 0.89) \\
Attention: Clinical + Macronutr. + Microbiome & \textbf{0.80} & $1.56{\times}10^{-14}$ & (0.71, 0.89) \\

\midrule

\multicolumn{4}{l}{\textit{Image-estimated macronutrients, CGMacros dataset}} \\

Berry: Clinical + Macronutrients & 0.41 & $1.09{\times}10^{-3}$ & (0.21, 0.64) \\
Berry: Clinical + Macronutrients + Microbiome & 0.39 & $2.48{\times}10^{-3}$ & (0.17, 0.63) \\

\addlinespace[0.4mm]

Qwen3.5-4B: Clinical + Macronutrients & 0.39 & $2.20{\times}10^{-3}$ & (0.10, 0.58) \\
Qwen3.5-4B: Clin. + Macronutr. + Microbiome & 0.16 & $2.32{\times}10^{-1}$ & (-0.20, 0.49)\\
Lingshu-7B: Clinical + Macronutrients & 0.03 & $8.44{\times}10^{-1}$ & (-0.22, 0.19) \\
Lingshu-7B: Clin. + Macronutr. + Microbiome & -0.01 & $9.64{\times}10^{-1}$ & (-0.17, -0.17) \\

\addlinespace[0.4mm]

Joint MLP: Clinical + Macronutrients & 0.49 & $7.20{\times}10^{-5}$ & (0.29, 0.75) \\
Joint MLP: Clinical + Macronutr. + Microbiome & 0.70 & $6.37{\times}10^{-10}$ & (0.42, 0.84) \\
Joint Attention: Clinical + Macronutrients & 0.59 & $8.77{\times}10^{-7}$ & (0.38, 0.86) \\
Joint Attention: Clin. + Macronutr. + Microbiome & \textbf{0.74} & $1.85{\times}10^{-11}$ & (0.58, 0.88) \\

\bottomrule
\end{tabular}
\label{tab:microbiome_comparison}
\end{table}
\begin{table}[t]
\centering
\caption{Ablation study evaluating glucose prediction without explicit macronutrient features.}
\small
\setlength{\tabcolsep}{4pt}
\begin{tabular}{lccc}
\toprule
\textbf{Model} & \textbf{R} & \textbf{p-value} & \textbf{95\% CI} \\
\midrule

\multicolumn{4}{l}{\textit{MLP glucose prediction head (no macronutrients)}}\\
\quad Clinical only & 0.34 & $8.18{\times}10^{-3}$ & (0.07, 0.65) \\
\quad Clinical + microbiome & 0.53 & $1.43{\times}10^{-5}$ & (0.05, 0.71) \\

\midrule

\multicolumn{4}{l}{\textit{Attention glucose prediction head (no macronutrients)}}\\
\quad Clinical only & 0.24 & $6.49{\times}10^{-2}$ & (-0.04, 0.71) \\
\quad Clinical + microbiome & 0.36 & $4.48{\times}10^{-3}$ & (0.07, 0.68) \\

\bottomrule
\end{tabular}
\label{tab:microbiome_comparison_ablation}
\end{table}

\subsection{Dataset and Preprocessing}

We use the CGMacros dataset~\cite{gutierrez-osuna_cgmacros_nodate} (2025), the first public dataset combining meal photographs, macronutrient composition, gut microbiome profiles (1980 binary bacterial features), clinical blood markers, continuous glucose monitoring, and anthropometrics for 45 adults (healthy, prediabetic, or T2D; 18--69 years) over 10 free-living days. To improve image quality, foil-wrapped meals were removed using few-shot CLIP detection, and only lunch meals were retained, yielding 363 meals after excluding missing values (\textbf{train}: 248/31 subjects, \textbf{val}: 56/7, \textbf{test}: 59/7). To reduce sparsity, microbiome features were aggregated at the family level~\cite{qian_guide_2020}, retaining the five most prevalent families. Continuous clinical and microbiome features were standardized using training-set statistics.

\subsection{Implementation}

Models were optimized using Adam with an initial learning rate of $5\times10^{-5}$ and weight decay of $5\times10^{-4}$. Separate learning-rate schedules were used for the image encoder (exponential decay, $\gamma=0.99$/epoch) and glucose prediction head (ReduceLROnPlateau, factor 0.5, patience 15, minimum learning rate $10^{-6}$). The objective combines Smooth L1 losses ($\beta=1$) on $\log(\mathrm{iAUC120})$ and $\log(\mathrm{AUC120})$ with a scale-normalized absolute-error loss for macronutrient prediction weighted by $\lambda_{\mathrm{macro}}=0.6$. Macronutrient and clinical features are embedded into tokens of dimension e=8. Dropout with probability p=0.2 is applied in the attention, FiLM, and prediction modules.

\subsection{Macronutrient Estimation}

Our replicated RGB-D model closely matches Shao et al.~\cite{shao_vision-based_2023} on Nutrition5K (20.85\% vs.\ 20.48\% mean PMAE). Performance decreases to 26.66\% on CGMacros lunch images, indicating a domain gap between controlled and free-living settings, but remains comparable to the RGB-only Nutrition5K baseline~\cite{thames_nutrition5k_2021} (30.43\%). This degradation likely reflects the difficulty of estimating portion size without reliable depth or mass cues in cluttered free-living images~\cite{thames_nutrition5k_2021,han_dpf-nutrition_2023,shao_vision-based_2023}. Nevertheless, the image-derived macronutrient estimates remain sufficiently informative to support downstream PPGR prediction.

\subsection{Baselines and Ablation Models}

As an architectural ablation, we replace the proposed attention-FiLM fusion module with a two-layer MLP (128 ReLU units per layer) operating on concatenated image-derived macronutrient, clinical, and microbiome features. No dropout or normalization is used. We additionally compare against existing PPGR baselines and general-purpose vision-language models (VLMs). Oracle experiments use ground-truth macronutrients, whereas image-based experiments use image-estimated macronutrients. "Joint" denotes end-to-end training of the image encoder and glucose prediction head.

\subsection{Glucose Prediction and Multimodal Fusion}

Table~\ref{tab:microbiome_comparison} compares the proposed attention-based model with the MLP ablation, existing PPGR baselines, and general-purpose vision-language models (VLMs) using both ground-truth (oracle) and image-estimated macronutrients.

The proposed attention-based model consistently outperforms the MLP ablation, while microbiome features improve prediction for both architectures. The best oracle model achieves $R=0.80$, and the best joint image-based model reaches $R=0.74$, surpassing Zeevi et al.~\cite{zeevi_personalized_2015} ($R=0.68$) and approaching Berry et al.~\cite{berry_human_2020} ($R=0.77$) despite using automatically estimated macronutrients and a substantially smaller free-living cohort.

To quantify the contribution of meal composition, Table~\ref{tab:microbiome_comparison_ablation} reports models without explicit macronutrient features. Performance drops substantially for both architectures, indicating that meal composition remains the dominant predictor of PPGR, while microbiome features provide complementary information.

The smaller performance gap between the attention-based and MLP models with image-estimated macronutrients (0.74 vs.\ 0.70) than with ground-truth macronutrients (0.80 vs.\ 0.72) suggests that errors in macronutrient estimation become the primary bottleneck once nutritional inputs are predicted rather than measured.

Compared with the Berry baseline, the proposed models remain more robust when using image-derived macronutrients, maintaining strong performance under noisy nutritional inputs. In contrast, general-purpose VLMs perform poorly, suggesting that semantic food representations alone are insufficient for individualized PPGR prediction without explicit modeling of nutritional and physiological factors.

\subsection{Limitations}

Our study is limited by both the dataset and the imaging modality. Free-living food photographs are inherently challenging, as portion size cannot be reliably inferred from RGB images and clutter, occlusion, and partially consumed meals further degrade macronutrient estimation. This likely explains much of the gap to the oracle model using ground-truth macronutrients, making more robust portion-size estimation an important direction for future work.

In addition, evaluation was performed on a relatively small number of independent test subjects. Although multiple meals were available per participant, patient-wise splitting reduces the effective sample size, resulting in relatively wide confidence intervals. Larger multimodal cohorts will be important for validating the proposed framework and improving the precision of performance estimates.
\section{Conclusion}

We presented a multimodal framework for personalized postprandial glycemic response (PPGR) prediction that integrates image-derived macronutrient estimates with clinical variables and gut microbiome information, eliminating the need for manually reported dietary intake. Our results demonstrate that image-derived nutritional representations provide a practical alternative to manual food logging, achieving competitive prediction performance under free-living conditions.

Integrating dietary, clinical, and microbiome information consistently improved prediction performance, highlighting the complementary value of these modalities for personalized glycemic modeling. Future work will focus on improving visual dietary estimation, particularly portion-size prediction, and validating the proposed framework on larger and more diverse multimodal cohorts.

\bibliographystyle{splncs04}
\bibliography{literature_bibtex}

\end{document}